\documentclass[10pt,twocolumn,letterpaper]{article}

\usepackage{cvpr}
\usepackage{times}
\usepackage{epsfig}
\usepackage{graphicx}
\usepackage{amsmath}
\usepackage{amssymb}
\usepackage{multirow}
\usepackage{comment}

\DeclareMathOperator*{\argmin}{argmin}

\usepackage[breaklinks=true,bookmarks=false]{hyperref}

\cvprfinalcopy 

\def\cvprPaperID{8863} 

\begin{document}

\title{Weakly Supervised Visual Semantic Parsing}

\author{Alireza Zareian, Svebor Karaman, and Shih-Fu Chang\\
Columbia University, New York, NY, USA\\
{\tt\small \{az2407,sk4089,sc250\}@columbia.edu
}}

\maketitle

\begin{abstract}

Scene Graph Generation (SGG) aims to extract entities, predicates and their semantic structure from images, enabling deep understanding of visual content, with many applications such as visual reasoning and image retrieval. Nevertheless, existing SGG methods require millions of manually annotated bounding boxes for training, and are computationally inefficient, as they exhaustively process all pairs of object proposals to detect predicates. In this paper, we address those two limitations by first proposing a generalized formulation of SGG, namely Visual Semantic Parsing, which disentangles entity and predicate recognition, and enables sub-quadratic performance. Then we propose the Visual Semantic Parsing Network, \textsc{VSPNet}, based on a dynamic, attention-based, bipartite message passing framework that jointly infers graph nodes and edges through an iterative process.
Additionally, we propose the first graph-based weakly supervised learning framework, based on a novel graph alignment algorithm, which enables training without bounding box annotations. Through extensive experiments, we show that \textsc{VSPNet} outperforms weakly supervised baselines significantly and approaches fully supervised performance, while being several times faster. We publicly release the source code of our method\footnote{\url{https://github.com/alirezazareian/vspnet}}.

\end{abstract}

\section{Introduction}

\begin{figure}[t]
\begin{center}
 \includegraphics[width=0.99\linewidth]{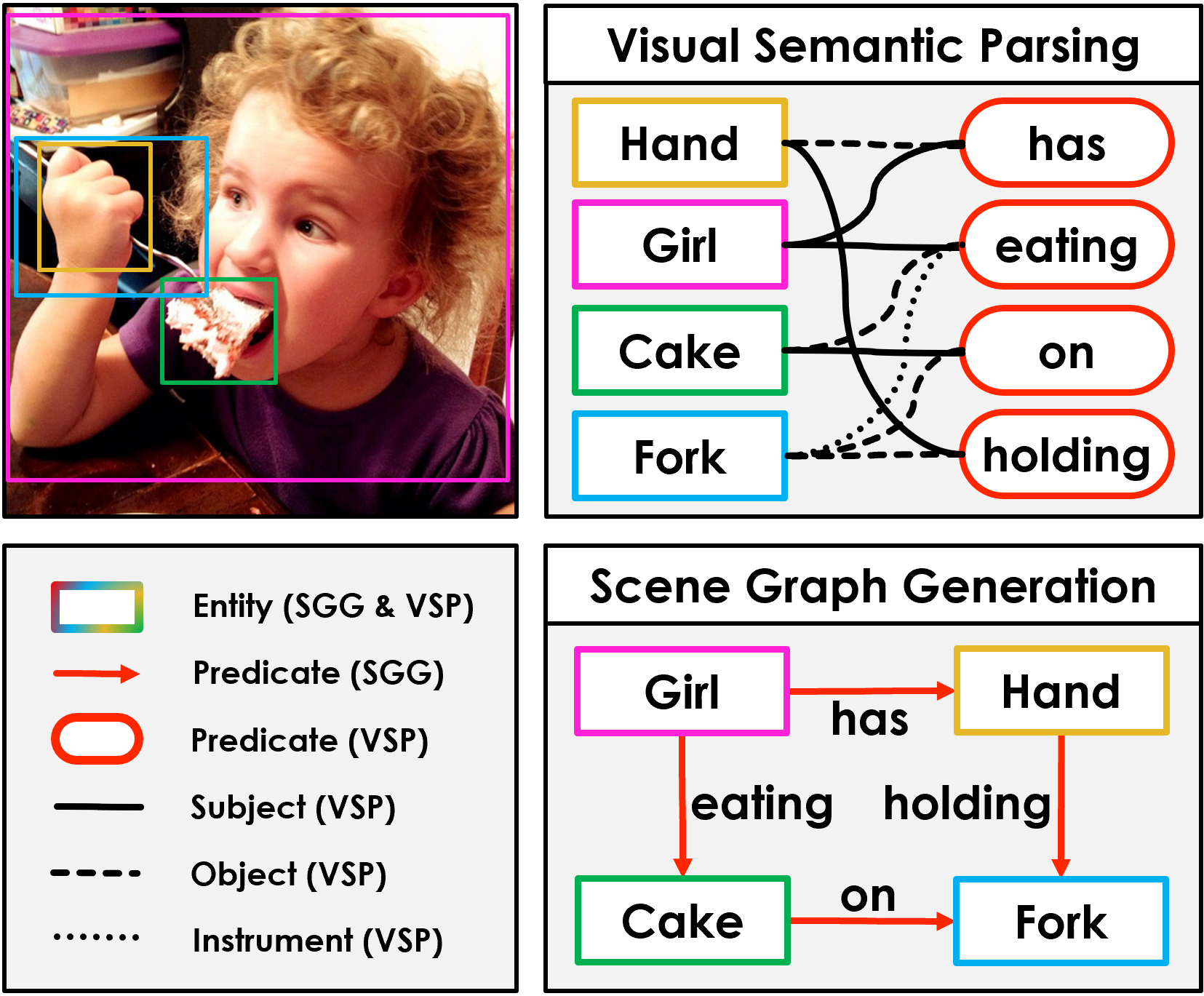}
\end{center}
\vspace{-0.1cm}
   \caption{An example of structured scene understanding formulated as Scene Graph Generation, where predicates are edges, compared to the proposed Visual Semantic Parsing, where predicates are nodes and edges represent semantic roles.}
\label{fig:vg_example}
\vspace{-0.3cm}
\end{figure}

Deep learning has excelled in various tasks such as object detection~\cite{ren2015faster} and speech recognition~\cite{bahdanau2016end}, but it falls short of tasks that require deeper semantic understanding and reasoning, such as Visual Question Answering (VQA)~\cite{zellers2018recognition}. 
Motivated by the success of structured representations in natural language processing \cite{banarescu2013abstract,sharma2015towards,wadden2019entity}, computer vision has started to adopt \textit{scene graphs} to improve performance and explainability, in various tasks such as VQA~\cite{shi2018explainable,hudson2019learning}, image captioning~\cite{yang2019auto}, and image retrieval~\cite{johnson2015image}. 
The task of Scene Graph Generation (SGG)~\cite{xu2017scene} aims to represent an image with a set of entities (nodes) and predicates (directed edges), as illustrated in
Figure~\ref{fig:vg_example} (bottom). Several methods have been proposed to address this problem~\cite{xu2017scene,li2017scene,yang2018graph,zellers2018neural}, but despite their success, important challenges remain unaddressed. 

Most existing methods are computationally inefficient, as they exhaustively process every pair of object proposals, in order to detect predicates. This results in a quadratic order with respect to the number of proposals. 
Extending to higher-order interactions 
has not been studied, and would make this problem even more complex. 
Furthermore, existing SGG methods require bounding box annotation for each object (node) in ground truth graphs, over the entire training data, which is an expensive constraint. 
We argue that SGG should ideally be disentangled from bounding box localization, so it can focus on high-level semantic and relational reasoning rather than low-level boundary analysis.
However, weakly supervised SGG has barely been studied, and the performance is far from supervised methods~\cite{zhang2017ppr}.

To advance structured scene understanding, we propose the Visual Semantic Parsing Network (\textsc{VSPNet}), which aims to address the two mentioned limitations, \ie, computation and supervision costs. 
To this end, we generalize the formulation of SGG to represent predicates as nodes in the same semantic space as entity nodes, and instead, represent \emph{semantic roles} (\eg subject and object) as edges. 
Figure~\ref{fig:vg_example} (top) illustrates the proposed \textit{Visual Semantic Parsing} (VSP) formalism. 
This not only allows us to break the quadratic complexity, but also can support higher-order interactions that cannot be expressed using the existing SGG formulation. For instance, the semantic structure of a \texttt{girl} \texttt{eating} \texttt{cake} \textit{with} \texttt{fork} can be represented as a predicate node, \texttt{eating}, connected to three entity nodes \texttt{girl}, \texttt{cake} and \texttt{fork}, via three types of edges that are labeled with \textit{subject}, \textit{object} and \textit{instrument} roles respectively.

Based on this new VSP formulation, we propose a dynamic, attention-based, bipartite message passing framework, which jointly infers node labels and edge labels through an iterative process, resulting in a VSP graph, and in turn a scene graph.
\textsc{VSPNet} consists of a \emph{role-driven} attention mechanism to dynamically estimate graph edges, along with a novel three-stage message aggregation network to route messages efficiently throughout the graph. These two modules successively refine nodes and edges of the graph, enabling a joint inference through global reasoning.
The proposed architecture does not need to process all pairs of object proposals and hence is computationally efficient. 
Finally and most importantly, we propose a novel framework to train \textsc{VSPNet} in weakly supervised settings, by defining a two-stage optimization problem and devising a novel graph alignment algorithm to solve it.

Through extensive experiments on the Visual Genome dataset,
we show that our method achieves significantly higher accuracy compared to weakly supervised counterparts, approaching fully supervised baselines. We also show that \textsc{VSPNet} is easily extendable to the fully supervised setting, where it can utilize bounding box annotations to further improve performance, and outperform the state of the art. Moreover, we show that our method is several times faster than all baselines, and qualitatively demonstrate its ability to extract higher-order interactions, which are beyond the capability of any existing method.

\section{Related work}

\noindent\textbf{Structured scene understanding:}
Deep learning often simplifies computer vision into classification or detection tasks that aim to extract visual concepts such as objects or actions in isolation. Lu \etal~\cite{lu2016visual} took a key step forward
by defining Visual Relationship Detection (VRD)~\cite{zhang2017visual, zhang2017ppr, liang2017deep, dai2017detecting, plummer2017phrase, yu2017visual, zhuang2017towards, jae2018tensorize}, which aims to classify relationships between pairs of objects detected in a scene. 
Their definition of ``relationship'', also known as \textit{predicate}, 
includes verbs (\eg \texttt{eating}), spatial positions (\eg \texttt{above}), and comparative adjectives (\eg \texttt{taller than}). 
Human-Object Interaction (HOI) detection~\cite{gkioxari2018detecting, chao2018learning, kato2018compositional, qi2018learning} is a specialized version of VRD that focuses on verbs with a human subject. 
More recently, Xu \etal~\cite{xu2017scene} redefined VRD as Scene Graph Generation (SGG)~\cite{li2017scene,newell2017pixels,zellers2018neural,yang2018graph,li2018factorizable,woo2018linknet}, which aims to jointly detect all objects and predicates in a scene, and represent it as a graph that captures the holistic scene content. 
SGG assumes exactly two entities (subject and object) involved in each predicate, which is not always the case in the real world.
Situation Recognition (SR)~\cite{yatskar2016situation, yatskar2017commonly, mallya2017recurrent} resolves that limitation by detecting a verb and all of its arguments in a scene, but does not localize the objects, and is limited to one verb per image. 
Our proposed VSP can be seen as a generalization of both SGG and SR, representing images with semantic graphs that could contain any number of predicates, localized entities, and semantic roles.

\noindent\textbf{Scene graph generation:}
The majority of SGG methods start by extracting object proposals from the input image, perform some kind of information propagation (\eg Bi-LSTMs in~\cite{zellers2018neural} or Graph Convolutional Nets in~\cite{yang2018graph}) to incorporate context, and then classify each proposal to an entity class, as well as each pair of proposals to a predicate class \cite{xu2017scene, li2017scene, zellers2018neural, li2018factorizable, woo2018linknet}. 
This process has a quadratic order and is thus inefficient. 
Recent methods have tried to reduce the computation by pruning the fully connected graph using a light-weight model~\cite{yang2018graph}, or by  factorizing the graph into smaller sub-graphs~\cite{li2018factorizable}.
However, they still suffer from quadratic order.
Newell and Deng \cite{newell2017pixels} proposed a method that does not rely on proposals at all, and directly extracts entities and predicates from a pair of feature maps. Our method is similar in that we allocate a constant, sub-quadratic number of predicates and infer their connection to entities, rather than processing all pairs of entities. 
In contrast with \cite{newell2017pixels} though, we base our graph on object proposals and exploit message passing to incorporate context.

\noindent\textbf{Neural message passing:}
Recent deep learning methods have increasingly utilized Message Passing (MP) in various computer vision tasks \cite{liu2018structure,chen2018iterative,kato2018compositional}. 
Most SGG methods use MP to propagate information among object proposals~\cite{xu2017scene, li2017scene, li2018factorizable, yang2018graph}. 
Instead of relying on a static, often fully-connected graph, we propose a dynamic, bipartite graph that is refined using attention to route messages between relevant entity-predicate pairs.  
In contrast with other dynamic MP methods that refine graph edges in each step, which have been used in other tasks such as HOI~\cite{qi2018learning} and video object detection~\cite{yuan2017temporal}, we define edges between entities and predicates rather than pairs of entities, leading to computational efficiency,
while incorporating the rich semantic role structure through three-stage aggregation.

\noindent\textbf{Weakly supervised learning:}
Weak Supervision (WS) has been advocated in several areas, such as object, action, and relation detection~\cite{bilen2016weakly,shou2018autoloc,zhang2017ppr}, and is motivated by the fact that manual annotation of boundaries is time consuming.
Most WS object detection methods are based upon multiple instance learning~\cite{dietterich1997solving}, which assumes each ground truth object corresponds to one out of many proposals, but the correspondence is unknown. WSDDN~\cite{bilen2016weakly} dedicates a network branch to select a proposal for each ground truth. Zhang \etal~\cite{zhang2017ppr} adopted WSDDN for VRD, selecting a pair of proposals for each ground truth relation. In contrast, we define a global optimization problem where the entire output graph has to be aligned with the ground truth graph, rather than considering each predicate independently. Peyre \etal~\cite{peyre2017weakly} defined a global optimization for WS VRD too, but it is limited to a linear regression model for relationship recognition. Our novel WS formulation allows learning with gradient descent, which enables us to train a deep network with a complex message passing architecture.
\section{Method}

\begin{figure*}[th]
\begin{center}
\includegraphics[width=\linewidth]{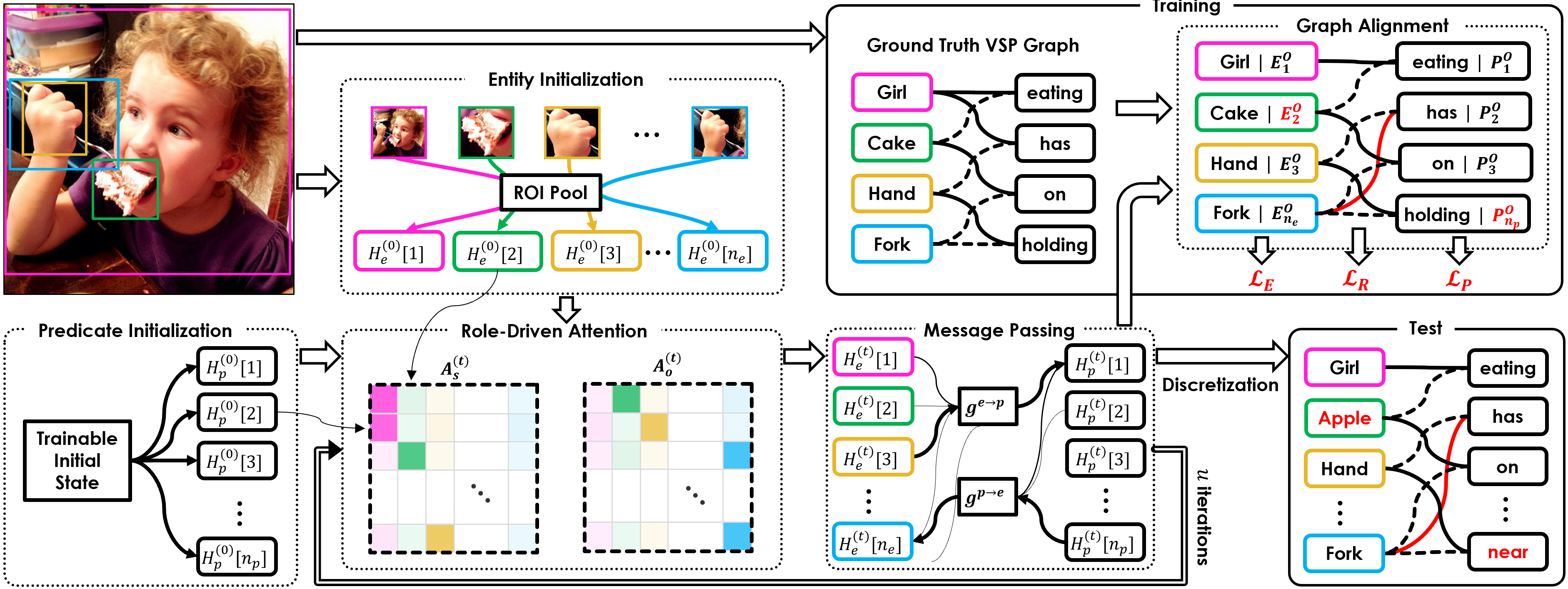}
\end{center}
\caption{Overview of our proposed framework: Given an input image and object proposals, a scene graph is produced by an iterative process involving a multi-headed attention module that infers edges between entities and predicates, and a novel message passing module to propagate information between nodes and update their states. 
To define a classification loss for each node and edge, the ground truth graph is aligned to our output graph through a novel weakly supervised algorithm. Red represents mistake. Best viewed in color.}
\label{fig:diagram}
\end{figure*}

In this section, we first formalize our problem in Section~\ref{sec:prob_form}, then detail our method and its two-fold contributions: the \textsc{VSPNet} architecture for constructing a semantic graph from an image (Section~\ref{sec:vspnet}), and a graph alignment algorithm for weakly supervised training of the proposed network (Section~\ref{sec:training}).
Figure~\ref{fig:diagram} illustrates the general pipeline of our method. 

\subsection{Problem formulation}
\label{sec:prob_form}

Given an image $I$, the goal of SGG is to produce a graph $G_\text{SGG}=(\mathcal{N}, \mathcal{E})$ where each node in $\mathcal{N}$ is represented by an entity class $c_i\in\mathcal{C}_e$ and a bounding box $b_i$, and each edge assigns a predicate class to an ordered pair of nodes, \ie, 
$\mathcal{E}: \mathcal{N} \times \mathcal{N} \mapsto \mathcal{C}_p$. 
The direction of predicate edges usually follow the order they would appear in an English phrase. For instance, a \texttt{person} \texttt{sitting on} \texttt{chair} would be represented as an edge labeled \texttt{sitting on}, going from the node \texttt{person} to the node \texttt{chair}, not the other way.

Nevertheless, this notation is inherently limiting, as it restricts predicates to have exactly two arguments present in the scene. 
This constraint may be acceptable for relational predicates such as prepositions, but certainly not for verbs, which constitute an important group of predicates.
To relax this constraint, we 
follow~\cite{yatskar2016situation} to adopt the formulation of Semantic Role Labeling~\cite{palmer2010semantic}, where predicates are represented as nodes, and edges represent semantic roles that entities play in each predicate. 
Accordingly, we define Visual Semantic Parsing (VSP) as predicting a bipartite graph $G_\text{VSP}=(\mathcal{N}_e, \mathcal{N}_p, \mathcal{E})$, where 
\begin{align}
\begin{split}
\label{eq:graph_def}
&\mathcal{N}_e = \Big\{\big( c_i\in\mathcal{C}_e, b_i\in\mathbb{R}^4 \big)\Big\}_{i=1}^{n_e}, \\ &\mathcal{N}_p = \big\{c_k\in\mathcal{C}_p\big\}_{k=1}^{n_p}, \text{ and}\\ &\mathcal{E}: \mathcal{N}_p \times \mathcal{N}_e \mapsto \mathcal{C}_r. 
\end{split}
\end{align}
Every scene graph $G_\text{SGG}$ has an equivalent VSP graph $G_\text{VSP}$ where each predicate has exactly two roles, subject and object, meaning $\mathcal{C}_r=\{s,o\}$.
However, an arbitrary VSP graph does not necessarily map to a scene graph, as a predicate may connect to less or more than two entities, potentially involving other semantic roles such as instrument. 
Hence, VSP is a generalization of SGG. 

In this paper we employ the VSP formalism, not only because it covers a wider range of semantics, but also because it naturally leads to a more efficient model architecture. 
In order to consider all possible relationships, most existing methods process a fully connected graph with $n_e^2$ edges, where $n_e$ is usually the number of proposals which is typically 300. 
This is while more than 99\% of graphs in Visual Genome have less than 20 predicates, and the largest one has 53. 
VSP allows us to replace the $n_e^2$ edges with a constant number of predicate nodes $n_p$, far less than $n_e^2$. 

\subsection{Visual semantic parsing network}
\label{sec:vspnet}

We propose \textsc{VSPNet}, which takes an image as input and generates a VSP graph. To this end, we utilize an object proposal network to initialize a set of \textit{entity nodes}, and devise another module to initialize a set of \textit{predicate nodes}. The goal of \textsc{VSPNet} is to classify each entity and predicate node into entity and predicate classes including background, and classify each entity-predicate pair into predefined edge types (semantic roles) including no-edge. 
These are two co-dependent tasks as incorporating nodes would be helpful for edge classification and vice versa. But since both of them are unknown and to be determined, our model successively infers each given the other.

More specifically, \textsc{VSPNet} is based on a novel bipartite message passing framework that propagates information from entities to predicates and vice versa, through a \textit{role-driven} attention mechanism that estimates edges. 
After nodes are updated using the estimated edges, we update edges by recomputing the attention using the new node representations, and repeat this process for $u$ iterations.
To incorporate each semantic role separately, we designate an attention head for each role.
This leads to a complex routing problem where messages from a potentially large number of nodes have to be propagated through multiple types of edges. Accordingly, we propose a three-stage message aggregation network to efficiently route and collect relevant messages for updating each node.

Formally, we define $H_e^{(0)} \in \mathbb{R}^{n_e \times d_e}$ to be the initial hidden state of $n_e$ entity nodes, and initialize each row using the appearance (RoI \cite{ren2015faster}) features of the corresponding object proposal, as well as its bounding box coordinates, by feeding them into two fully connected networks $e_a(.)$ and $e_b(.)$, and adding the two outputs. We also define $H_p^{(0)} \in \mathbb{R}^{n_p \times d_p}$ to be the initial hidden state of $n_p$ predicate nodes. 
$H_p^{(0)}$ is a trainable matrix, randomly initialized before training but fixed during test.
Given $H_e^{(t)}$ and $H_p^{(t)}$, we compute a set of attention matrices $\tilde{A}_r^{(t)} \in \mathbb{R}^{n_p \times n_e}$, each representing a semantic role class $r$ in $\mathcal{C}_r$: 
\begin{align}
\begin{split}
\tilde{A}_r^{(t)}[k,i] = \Big\langle f_r^p\big(H_p^{(t)}[k]\big) \;,\; f_r^e\big(H_e^{(t)}[i]\big) \Big\rangle,
\end{split}
\end{align}
where
$\langle.,.\rangle$ represents dot product, $H[k]$ represents the $k$th row of $H$, and $f_r^p$ and $f_r^e$ are trainable fully connected networks to compute the query and key vectors of the attention. 
We further stack $\tilde{A}_r^{(t)}$ to build the 3-dimensional tensor $\tilde{A}^{(t)}$ that represents the entire role-driven attention. 
In our experiments, no predicate can take more than one entity for each role, and no entity-predicate pair can have more than one semantic role. Hence, we normalize $\tilde{A}^{(t)}$ such that:
\begin{align}
\begin{split}
A_r^{(t)}[k,i] &= \frac{\exp\big(\tilde{A}_r^{(t)}[k,i]\big)}{p_\varnothing + \sum_{r'=1}^{n_r} \exp\big(\tilde{A}_{r'}^{(t)}[k,i]\big)} \\
&\times \frac{\exp\big(\tilde{A}_r^{(t)}[k,i]\big)}{p_\varnothing + \sum_{i'=1}^{n_e} \exp\big(\tilde{A}_r^{(t)}[k,i']\big)}.
\end{split}
\end{align}
This can be interpreted as applying two softmax functions in paralell on $\tilde{A}^{(t)}$, once normalizing along the axis of roles, and once along the axis of entities, and then multiplying the two normalized matrices, element-wise. The constant $p_\varnothing$ is added to each denominator to allow the sum to be less than one, \eg no role between an entity-predicate pair.

After computing attention matrices, we use them to propagate information from each entity to its relevant predicates and vice versa. To this end, we propose a three-stage message aggregation framework, that computes the incoming message to update each node, by aggregating outgoing messages from all other nodes, and separately processing them in the context of each semantic role. More specifically: 
\begin{align}
\begin{split}
& M_p^{(t)}[k] = g^{e \rightarrow p} \big( A^{(t)}, H_e^{(t)} \big)
\\&= g^{p\leftarrow}\Bigg( \sum_{r=1}^{n_r} g_r^{e} \Big( \sum_{i=1}^{n_e}  A_r^{(t)}[k,i] g^{e\rightarrow} \big( H_e^{(t)}[i] \big)\Big)\Bigg),
\end{split}
\end{align}
where $g_r^{e\rightarrow}$, $g_r^e$, and $g_r^{p\leftarrow}$ are independent, trainable fully connected networks, respectively called \textit{send head}, \textit{pool head}, and \textit{receive head}. Note that the pool head consists of $n_r$ separate networks applied on the pooled messages for each role. Similarly, the incoming message to update each entity is computed as:
\begin{align}
\begin{split}
& M_e^{(t)}[i] = g^{p \rightarrow e} \big( A^{(t)}, H_p^{(t)} \big)
\\&= g^{e\leftarrow}\Bigg( \sum_{r=1}^{n_r} g_r^{p} \Big( \sum_{k=1}^{n_p}  A_r^{(t)}[k,i] g^{p\rightarrow} \big( H_p^{(t)}[k] \big)\Big)\Bigg).
\end{split}
\end{align}
After collecting messages for each node, we update their state using two Gated Recurrent Units (GRU) \cite{cho2014learning}.
\begin{align}
\begin{split}
&H_e^{(t+1)}[i] = \text{GRU}_e\Big( H_e^{(t)}[i], M_e^{(t)}[i] \Big) \text{, and}\\
&H_p^{(t+1)}[k] = \text{GRU}_p\Big( H_p^{(t)}[k], M_p^{(t)}[k] \Big) .
\end{split}
\end{align}
This process is repeated for a constant number of times $u$, and the final states $H_e^{(u)}$ and $H_p^{(u)}$ are passed through another pair of fully connected networks ($h_e$, $h_p$) to produce semantic embeddings $E^O$ and $P^O$ for entity and predicate nodes. The final state of the adjacency matrices $A_r^{(u)}$ are stacked together and named $A^O$. 

After the message passing process, we have a continuous and fully differentiable output graph $G^O_\text{soft} = (E^O, P^O, A^O)$. 
In order to produce a valid, discrete graph as defined in Eq.~(\ref{eq:graph_def}), 
we apply a two-step \textit{discretization} process.
First, we convert $E^O$ and $P^O$ to discrete labels by picking the nearest neighbor of each of their rows among a dictionary of entity and predicate class embeddings. 
Next, we threshold the attention matrix $A^O$ and suppress non-maximum roles for each entity-predicate pair.
This leads to a discrete graph $G^O = (\mathcal{N}_e^O, \mathcal{N}_p^O, \mathcal{E}^O)$.
In the next subsection, we define our cost function, where we also need the opposite process: converting a ground truth graph $G^T = (\mathcal{N}_e^T, \mathcal{N}_p^T, \mathcal{E}^T)$ to a soft representation $G^T_\text{soft} = (E^T, P^T, A^T)$. 
To this end, we stack the class embedding of entity and predicate nodes to get matrices $E^T$ and $P^T$, and encode the edges into a binary adjacency matrix $A^T$.

\subsection{Weakly supervised training}
\label{sec:training}

We train our model using pairs of image and unlocalized ground truth graph.
Specifically, we need to compare the soft output graph $G^O_\text{soft}$ (\ie before discretization) to the target $G^T_\text{soft}$ to calculate a differentiable cost to be minimized. 
To this end, we find an alignment (\ie, node correspondence) between the 
two
graphs, and then define the overall cost as a summation of loss terms over aligned nodes and edges.  
Formally, we define an alignment $\mathcal{I}$ as: 
\begin{align}
\begin{split}
&\mathcal{I} = (\mathcal{I}_e, \mathcal{I}_p) 
\text{, where}\\
&\mathcal{I}_e = \Big\{ (i,j) | i\in\{1...n_e^O\}, j\in\{1...n_e^T\} \Big\}
\text{, and}\\
&\mathcal{I}_p = \Big\{ (k,l) | k\in\{1...n_p^O\}, l\in\{1...n_p^T\} \Big\},
\end{split}
\end{align}
where $n_e^O=n_e$ and $n_p^O=n_p$ are the number of output entity and predicate nodes, while $n_e^T$ and $n_p^T$ are the number of ground truth entity and predicate nodes. $\mathcal{I}_e$ is a valid entity alignment if for any output node $i$ there is at most one target node $j$, and for each $j$ there is at most one $i$, where $(i,j) \in \mathcal{I}_e$. A similar constraint holds for $\mathcal{I}_p$.
Moreover, $\mathcal{I}_e$ is a \textit{maximal alignment} if all output entities \textbf{or} all target entities are aligned, whichever is fewer, \ie
\begin{align}
\begin{split}
\label{eq:opt_sum_cons}
&\lvert \mathcal{I}_e \rvert = \min(n_e^O, n_e^T)
,\;\text{and similarly,}\\
&\lvert \mathcal{I}_p \rvert = \min(n_p^O, n_p^T),
\end{split}
\end{align}
where $\lvert . \rvert$ denotes set cardinality.
Given an alignment $\mathcal{I}$ between output and target graphs, our objective function is: 
\begin{align}
\begin{split}
\label{ex:overal_loss}
\mathcal{L}(G^O, G^T, \mathcal{I}) = \mathcal{L}_E + \mathcal{L}_P + \lambda \mathcal{L}_R,
\end{split}
\end{align}
which is a combination of costs for entity recognition, predicate recognition, and semantic role labeling. 

Our weakly supervised training framework is independent of how we define each loss term, as long as they are a summation of costs over aligned nodes. For instance, if we define the entity loss $\mathcal{L}_E$ and predicate loss $\mathcal{L}_P$ as mean square errors of entity and predicate embeddings, and if we define the role loss $\mathcal{L}_R$ to be a binary cross entropy on all attention scores, we can write:
\begin{align}
\label{eq:ent_loss}
\begin{split}
\mathcal{L}_E(G^O, G^T, \mathcal{I}) =
\frac{1}{\lvert \mathcal{I}_e \rvert} \sum_{(i,j) \in \mathcal{I}_e} \left\lVert E_i^O - E_j^T \right\rVert_2^2,
\end{split}
\end{align}
\begin{align}
\begin{split}
\mathcal{L}_P(G^O, G^T, \mathcal{I}) = \frac{1}{\lvert \mathcal{I}_p \rvert} \sum_{(k,l) \in \mathcal{I}_p} \left\lVert P_k^O - P_l^T \right\rVert_2^2,
\end{split}
\end{align}
\begin{align}
\begin{split}
\mathcal{L}_R(G^O, G^T, \mathcal{I}) = 
\frac{1}{n_r} 
\sum_{r=1}^{n_r} 
\mathcal{\mathcal{L}}_r
,\end{split}
\end{align}
where for role $r$,
\begin{align}
\begin{split}
\mathcal{\mathcal{L}}_r=
\frac{1}{\lvert \mathcal{I} \rvert} 
\sum_{(i,j) \in \mathcal{I}_e} \sum_{(k,l) \in \mathcal{I}_p}
 \mathcal{X}\big(A_r^O[k, i], A_r^T[l, j]\big),
\end{split}
\end{align}
where $\lvert \mathcal{I} \rvert = \lvert \mathcal{I}_e \rvert \lvert \mathcal{I}_p \rvert$, and
\begin{align}
\begin{split}
\mathcal{X}(p,q) = - q \log p - (1-q) \log (1-p).
\end{split}
\end{align}
Since $\mathcal{L}_R$ is in a different scale than $\mathcal{L}_E$ and $\mathcal{L}_P$, we use a hyperparameter $\lambda$ to balance its significance in Eq.~(\ref{ex:overal_loss}).

The main challenge of weakly supervised learning is that the alignment $\mathcal{I}$ is not known, and thus our training involves the following nested optimization:
\begin{align}
\begin{split}
\label{eq:optimization}
\phi^* = \argmin_\phi \mathbb{E}\big[\min_\mathcal{I} \mathcal{L}(G^O, G^T, \mathcal{I})\big],
\end{split}
\end{align}
where $\phi$ is the collection of model parameters that lead to $G^O$, and the expectation is estimated by averaging over minibatches sampled from training data. Note that the inner optimization is subject to the constraints in Eq.~(\ref{eq:opt_sum_cons}). Inspired by the EM algorithm~\cite{moon1996expectation}, we device an alternating optimization approach: We use the Adam Optimizer \cite{kingma2014adam} for the outer optimization, and propose an iterative alignment algorithm to solve the inner optimization in the following.

\begin{table*}[tb]
\begin{center}
\begin{tabular}{l|c|c c|c c}
\hline
\multirow{2}{*}{Method} & \multirow{2}{*}{Supervision} & \multicolumn{2}{|c|}{\textsc{SGGen}} & \multicolumn{2}{|c}{\textsc{PhrDet}} \\
&& R@50 & R@100 & R@50 & R@100 \\ 
\hline
\hline
VtransE-MIL \cite{zhang2017ppr} & \multirow{2}{*}{Weak} & 0.7 & 0.9 & 1.5 & 2.0  \\

PPR-FCN \cite{zhang2017ppr} &  & 1.5 & 1.9 & 2.4 & 3.2  \\

\hline

\textsc{VSPNet} w/o iterative alignment & \multirow{6}{*}{Weak} & 1.3 & 1.6 & 8.0 & 10.2 \\
\textsc{VSPNet} w/ fewer alignment steps &  & 1.8 & 2.0 & 9.9 & 11.9 \\
\textsc{VSPNet} w/o three-stage MP &  & 2.4 & 2.8 & 16.7 & 19.8 \\
\textsc{VSPNet} w/o role-driven MP &  & 2.5 & 2.9 & 15.7 & 18.7 \\
\textsc{VSPNet} w/ fewer MP steps &  & 2.5 & 2.8 & 15.5 & 18.3 \\

\textsc{VSPNet} (Ours) &  & \textbf{3.1} & \textbf{3.5} & \textbf{17.6} & \textbf{20.4}  \\
\hline
VtransE \cite{zhang2017ppr} & \multirow{3}{*}{Full} & 5.5 & 6.0 & 9.5 & 10.4 \\

S-PPR-FCN \cite{zhang2017ppr} &  & 6.0 & 6.9 & 10.6 & 11.1\\

\textsc{VSPNet} (Ours) &  & \textbf{8.9} & \textbf{9.9} & \textbf{24.0} & \textbf{27.8}  \\
\hline
\end{tabular}
\end{center}
\caption{Results on VG preprocessed by~\cite{zhang2017ppr}. All numbers are in percentage and baselines were borrowed from~\cite{zhang2017ppr}}
\label{table:vg_zhang}
\vspace{-0.2cm}
\end{table*}

\label{sec:alignment}
There are no efficient exact algorithms for solving the inner optimization in Eq.~(\ref{eq:optimization}). 
Hence, we propose an iterative algorithm to approximate the optimal alignment. 
We show that given an entity alignment $\mathcal{I}_e$, it is possible to find the optimal predicate alignment $\mathcal{I}_p$ in polynomial time, and similarly from $\mathcal{I}_p$ to $\mathcal{I}_e$. Accordingly, we perform those two steps iteratively in a coordinate-descent fashion, which is guaranteed to converge to a local optima. 

Supposing $\mathcal{I}_e$ is given, we intend to find $\mathcal{I}_p$ that minimizes $\mathcal{L}$. Since $\mathcal{L}_E$ is constant with respect of $\mathcal{I}_p$, the problem reduces to minimizing $\mathcal{L}_P + \lambda\mathcal{L}_R$, which can be written:
\begin{align}
\begin{split}
\label{eq:opt_side_1}
&\mathcal{L}_P + \lambda\mathcal{L}_R = 
\frac{1}{\lvert \mathcal{I}_p \rvert} \sum_{(k,l) \in \mathcal{I}_p} W^{P}_{k l},
\end{split}
\end{align}
where $W^{P}$ is a pairwise cost function between output and target predicate nodes, measuring not only their semantic embedding distance, but also the discrepancy of their connectivity in graph. More specifically:
\begin{align}
\begin{split}
W^{P}_{kl} \triangleq &
\left\lVert P_{k}^O - P_{l}^T \right\rVert_2^2 + 
\\& \;
\frac{\lambda}{n_r \lvert \mathcal{I}_e \rvert}
\sum_{(i,j) \in \mathcal{I}_e}
\sum_{r=1}^{n_r} 
\mathcal{X}\big(A_r^O[k, i], A_r^T[l, j]\big).
\end{split}
\end{align}
Note that the optimization of Eq.~(\ref{eq:opt_side_1}) is subject to Eq.~(\ref{eq:opt_sum_cons}), which makes $\lvert \mathcal{I}_p \rvert$ a constant. Hence, this problem is equivalent to maximum bipartite matching with fully connected cost function $W^{P}$, which can be solved in polynomial time using the Kuhn-Munkres algorithm \cite{munkres1957algorithms}.

Similarly, given $\mathcal{I}_p$, we can solve for $\mathcal{I}_e$, and repeat alternation. 
Every step leads to a lower or equal loss since either $\mathcal{L}_P + \mathcal{L}_R$ is minimized while $\mathcal{L}_E$ is fixed, or $\mathcal{L}_E + \mathcal{L}_R$ is minimized while $\mathcal{L}_P$ is fixed. 
Since $\mathcal{L}$ cannot become negative, these iterations must converge. 
We have observed that the convergence value of $\mathcal{L}$ is not sensitive to whether we start by initializing $\mathcal{I}_e$ or $\mathcal{I}_p$, nor does it depend on the initialization value. 
In our experiments we initialize $\mathcal{I}_p$ to an empty set and proceed with updating $\mathcal{I}_e$.
We denote by $v$ the number of iterations used for this alignment procedure.

Our method can be naturally extended to the fully supervised setting by adding a term in Eq. \ref{eq:ent_loss}, to maximize the overlap between the aligned pairs of bounding boxes. 
Specifically, we redefine $\mathcal{L}_E$ as:
\begin{align}
\begin{split}
\mathcal{L}_E^\text{sup}(G^O, G^T, \mathcal{I}) &=
\frac{1}{\lvert \mathcal{I}_e \rvert} \sum_{(i,j) \in \mathcal{I}_e}  \Big( 
\left\lVert E_i^O - E_j^T \right\rVert_2^2
\\&- 
\lambda_B \log \big(\text{IoU}[B_i^O - B_j^T] + \epsilon\big) \Big)
,\end{split}
\end{align}
where $B^O$ and $B^T$ are the set of output and ground truth bounding boxes respectively, and $\lambda_B$ and $\epsilon$ are hyper-parameters selected by cross-validation. Note that the gradient of the added term with respect to model parameters is zero, and hence this only affects alignment.

\section{\label{sec:exp}Experiments}

We apply our framework on the Visual Genome (VG) dataset \cite{krishna2017visual} for the task of scene graph generation, and compare to both weakly and fully supervised baselines. Through quantitative analysis, we show that \textsc{VSPNet} significantly outperforms the weakly and fully supervised state of the art, while being several times faster than existing methods. Furthermore, ablation experiments show the contribution of each proposed module, namely iterative alignment, role-driven attention, and three-stage message aggregation.
We finally provide qualitative evidence that our method is able to produce VSP graphs, which are beyond the expressive capacity of conventional scene graphs.

\begin{table*}[tb]
\begin{center}
\begin{tabular}{l|c|c|c c|c c|c c}
\hline
\multirow{2}{5em}{Method} & \multirow{2}{5em}{Supervision} & \multicolumn{3}{|c|}{\textsc{SGGen}} & \multicolumn{2}{|c|}{\textsc{SGCls}} &
\multicolumn{2}{|c}{\textsc{PredCls}} 
\\
&& Time & R@50 & R@100 & R@50 & R@100 & R@50 & R@100  \\ 
\hline\hline
IMP~\cite{xu2017scene} & \multirow{7}{*}{Full} & 1.64 & 3.4 &4.2&21.7&24.4&44.7&53.1\\

MSDN~\cite{li2017scene} &  & 3.56 & 7.7 & 10.5 & 19.3 & 21.8 & 63.1 & 66.4\\

MotifNet~\cite{zellers2018neural} &  & 2.07 & 6.9 & 9.1 & 23.8 & 27.2 & 41.8 & 48.8\\

Assoc. Emb.~\cite{newell2017pixels} &  & 1.19 & 9.7 &11.3&26.5&30.0&\textbf{68.0}&\textbf{76.2}\\

Graph R-CNN~\cite{yang2018graph} &  & 0.83 & 11.4 & 13.7 & 29.6 & 31.6 & 54.2 & 59.1 \\


\textsc{VSPNet} (Ours) &  & \textbf{0.11} & \textbf{12.6} & \textbf{14.2} & \textbf{31.5} & \textbf{34.1} & 67.4 & 73.7\\
\hline
\textsc{VSPNet} (Ours) & Weak & \textbf{0.11} & \textbf{4.7} & \textbf{5.4} & \textbf{30.5} & \textbf{32.7} & \textbf{57.7} & \textbf{62.4}\\
\hline
\end{tabular}
\end{center}
\caption{Results on VG~\cite{xu2017scene}. Recall numbers (\%) are from~\cite{yang2018graph}. Inference time is in seconds per image, partially borrowed from~\cite{li2018factorizable}.  
}
\label{table:vg_xu}
\vspace{-0.2cm}
\end{table*}

\subsection{Implementation details}

We use an off-the-shelve Faster R-CNN~\cite{ren2015faster} 
pretrained on the Open Images dataset~\cite{kuznetsova2018open} to extract object proposals that are needed as inputs to \textsc{VSPNet}. We extract proposal coordinates and features once for all images, and keep them fixed while training and evaluating our model. We do not stack \textsc{VSPNet} on top of Faster R-CNN and do not fine-tune Faster R-CNN during training.
We use the original implementation of GRU \cite{cho2014learning} with 1024-dimensional states ($d_e$ and $d_p$).
The initialization heads $e_a$ and $e_b$, the attention heads $f_r^e$ and $f_r^p$, and the message passing heads, $g^{e\rightarrow}$, $g_r^e$, $g^{p\leftarrow}$, $g^{p\rightarrow}$, $g_r^p$, and $g^{e\leftarrow}$, are all fully connected networks with two 1024-dimensional layers. 
The embedding prediction heads $h_e$ and $h_p$ are each single-layer networks that map 1024-D GRU states to the 300-D embedding space.
All fully connected networks use leaky ReLU activation functions \cite{he2015delving}. 
Through cross-validation, we set $\lambda=10$, $u=3$, and $v=3$.
We use GloVe embeddings \cite{pennington2014glove} to represent each class, and we fine-tune it during training. 

The number of predicate nodes $n_p$ is an important choice. Having more predicate nodes will increase recall but also inference time. Since SGG methods are conventionally evaluated at 100 and 50 predicates, we set $n_p=100$. To output only 50 predicates, we rank the predicate nodes with respect to their confidence, which is defined as the product of three classification confidence scores, for subject, object and predicate.
To report inference time in Table~\ref{table:vg_xu}, we compute the average inference time per image on the test set, using identical settings for all methods (NVIDIA TITAN X, 200 proposals, VGG backbone). The time includes the extraction of proposals and their features.

\subsection{Task definition}

The Visual Genome dataset consists of 108,077 images with manual annotation of objects and relationships, with open-vocabulary classes. 
\cite{xu2017scene} and \cite{zhang2017ppr} preprocess the annotated objects and relationships to produce scene graphs with a fixed vocabulary. \cite{xu2017scene} keeps 150 most frequent entity and 50 most frequent predicate classes, while \cite{zhang2017ppr} cuts at 200 and 100 respectively. We perform two sets of experiments, based on both \cite{xu2017scene} and \cite{zhang2017ppr}, to be able to compare to the performances reported by each paper separately. We follow their preprocessing, data splits, and evaluation protocol, but we assume bounding boxes are not available during weakly supervised training.

The main evaluation metric dubbed \textsc{SGGen}, measures the accuracy of \texttt{subject}-\texttt{predicate}-\texttt{object} triplets. A detected triplet is considered correct if the predicted class for subject, object, and predicate are all correct, and the subject and object bounding boxes have an Intersection over Union (IoU) of at least 0.5 with ground truth. To evaluate, the top $K$ triplets predicted by the model are matched to ground truth triplets. The number of correctly matched triplets is divided by the total number of triplets in the ground truth to compute recall at $K$. This value is averaged over all images leading to R@50 and R@100.
Since \textsc{SGGen} is highly affected by the quality of object proposals.
we also report \textsc{SGCls}, which assumes ground truth bounding boxes are given at test time, instead of proposals. 
Another metric, \textsc{PredCls} assumes ground truth bounding are given, and true object classes are given too. \cite{zhang2017ppr} also evaluates using \textsc{PhrDet}, which stands for Phrase Detection. This metric is similar to \textsc{SGGen}, with the difference that instead of evaluating the bounding box of subject and object separately, the goal is to predict a union bounding box enclosing both the object and subject. To this end, for each detected triplet, we get the union box of its subject and object, and match with that of ground truth triplets at IoU~$\geq0.5$.

\subsection{Results}

Table~\ref{table:vg_zhang} shows our quantitative results on VG compared to VtransE~\cite{zhang2017visual} and PPR-FCN~\cite{zhang2017ppr}, in both Weakly Supervised (WS) and Fully Supervised (FS) settings, following the evaluation settings of~\cite{zhang2017ppr}. 
Our \textsc{VSPNet} achieves the best WS performance, 
with \textsc{SGGen} performance more than two times higher and \textsc{PhrDet} more than six times higher than the state of the art.
Moreover, the FS extension of our method outperforms the FS variants of those baselines significantly. On the \textsc{PhrDet} measure, even our WS method outperforms all FS baselines.
Furthermore, we provide ablative variants of our method as extra rows in Table~\ref{table:vg_zhang}, 
to study the effect of each proposed component in isolation.

In \textsc{VSPNet} \textbf{w/o iterative alignment}, we replace the proposed alignment algorithm with a heuristic baseline, where we align entities by minimizing $\mathcal{L}_E$ and independently align predicates to minimize $\mathcal{L}_P$, in a one-step process. 
Our alignment algorithm leads to more than twice the performance of this ablation. We make a similar observation by reducing the number of alignment steps $v$ from 3 to 1, denoted as \textsc{VSPNet} \textbf{w/ fewer alignment steps}.
Furthermore, in \textsc{VSPNet} \textbf{w/o three-stage MP}, we replace the proposed three-stage message aggregation framework with a conventional average pooling, that computes the sum of all messages after multiplying by the attention weights. In \textsc{VSPNet} \textbf{w/o role-driven MP}, we keep the three-stage message aggregation, but remove the role-driven attention, and replace $A_r(t)$ with a constant, uniformly distributed attention. 
Finally, in \textsc{VSPNet} \textbf{w/ fewer MP steps}, we only reduce the number of MP steps, $u$, from 3 to 1. All these three ablations lead to inferior performance, proving the effectiveness of our proposed message passing framework.

To compare to more recent methods, we also perform experiments on the original version of VG that was used by \cite{xu2017scene}, and follow the evaluation protocol of \cite{yang2018graph}. Table~\ref{table:vg_xu} compares \textsc{VSPNet} to all the numbers reported by \cite{yang2018graph}.
The FS version of our method outperforms all state-of-the-art methods in all metrics, except slightly outperformed by Assoc. Emb.~\cite{newell2017pixels} in \textsc{PredCls} only. 
In addition to superior accuracy, our method is several times faster than all methods. It is also 5 times faster than Factorizable Net~\cite{li2018factorizable}, which is the fastest SGG method (0.55 seconds per image), although not shown in Table~\ref{table:vg_xu}, because their reported recall is computed differently than ours.

Furthermore, our WS method shows competitive performance and even outperforms some FS methods. Although there is a performance drop from FS to WS, that is mainly due to the difficulty of object localization in the WS setting.  
In \textsc{SGCls}, it achieves a performance very close to FS \textsc{VSPNet}, and outperforms all other FS baselines. 
This suggests that if some day we have access to very accurate proposals, our WS model would perform as accurately as FS methods.
Note that although \textsc{SGCls} provides ground truth bounding boxes, the WS model 
only treats them as input proposals, and is still trained with unlocalized ground truth and unknown alignment. 
Also note that all baselines in Table~\ref{table:vg_xu} train their Faster R-CNN on VG directly, using annotated bounding boxes that we assume not available in WS settings. Hence, we use an off-the-shelve Faster R-CNN that is pretrained on another dataset in all our experiments. This makes the comparison in Table~\ref{table:vg_xu} somewhat unfair, to our disadvantage. Adopting the backbone used by the baselines would improve our results, but violates WS constraints.

\begin{figure}[]
\begin{center}
   \includegraphics[width=0.54\linewidth]{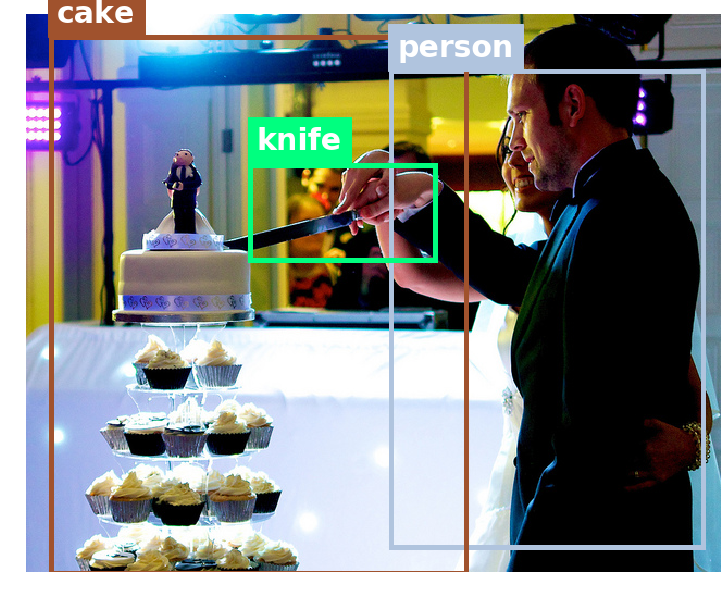}
   \includegraphics[width=0.45\linewidth]{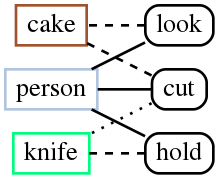} \\ \vspace{3mm}
   \includegraphics[width=0.54\linewidth]{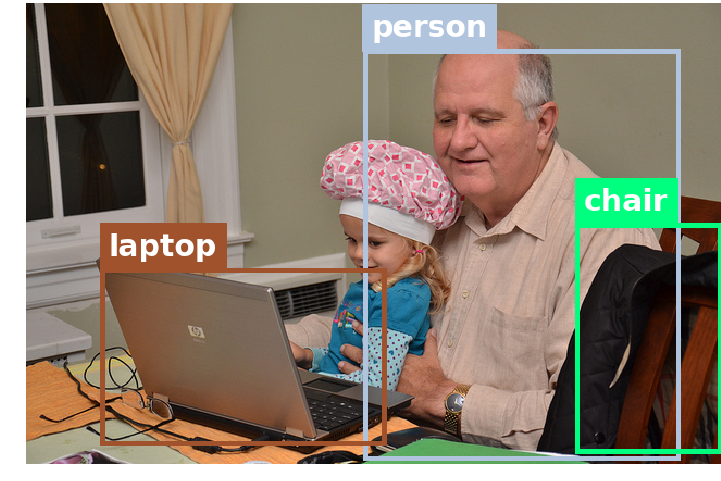}
   \includegraphics[width=0.45\linewidth]{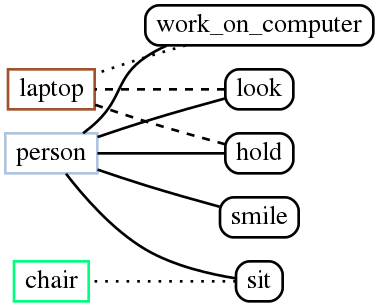} \\ \vspace{3mm}
   \includegraphics[width=0.54\linewidth]{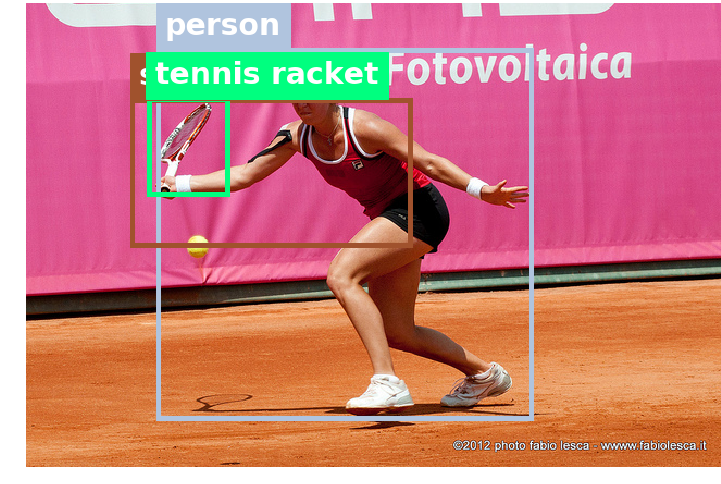}
   \includegraphics[width=0.45\linewidth]{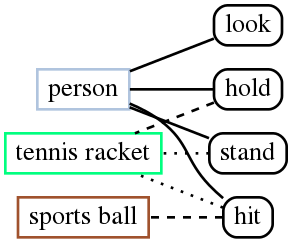}
\end{center}
   \caption{Example VSP graphs generated by our method. Solid, dashed, and dotted lines represent subject, object, and instrument. 
   }
\label{fig:examples}
\vspace{-0.3cm}
\end{figure}

To illustrate the expressive power of our novel VSP formulation, we train our model on the V-COCO dataset~\cite{gupta2015visual}, which annotates human actions in images, as well as objects and instruments of those actions. While this dataset has been primarily used for HOI in the literature~\cite{qi2018learning,wan2019pose}, we adopt it for VSP, by aggregating all action annotations of each image into a single semantic graph, and connecting them to the related objects through 3 types of semantic role: subject, object, and instrument. The resulting VSP graphs have unique properties that are not seen in scene graphs, as shown in Figure~\ref{fig:examples}, such as verbs with more than two entities (\eg \texttt{person} \texttt{cutting} \texttt{cake} \textit{with} \texttt{knife}), and verbs with only one entity (\eg \texttt{person} \texttt{smiling}). After training our model on the training set of V-COCO, we apply it on the test set and visualize output graphs in Figure~\ref{fig:examples}. Our method successfully generates VSP graphs containing interactions that are not possible with any SGG method. 
\section{Conclusion}

We proposed a method to parse an image into a semantic graph that includes entities, predicates, and semantic roles. 
Unlike prior works, our method does not require bounding box annotations for training, and does not rely on exhaustive processing of all object proposal pairs. Moreover, it is able to extract more flexible graphs where any number of entities are involved in each predicate.
To this end, we proposed a generalized formulation of Scene Graph Generation (SGG) that disentangles predicates from entities, and enables sub-quadratic performance. Based on that, we proposed \textsc{VSPNet}, based on a dynamic, attention-based, bipartite message passing framework. We also introduced the first graph-based weakly supervised learning framework based on a novel graph alignment algorithm.
We compared our method to the state of the art through extensive experiments, and achieved significant performance improvements in both weakly supervised and fully supervised settings, while several times faster than every existing method.

\paragraph{Acknowledgements:}
\small This work was supported by the U.S. DARPA AIDA Program No. FA8750-18-2-0014. The views and conclusions contained in this document are those of the authors and should not be interpreted as representing the official policies, either expressed or implied, of the U.S. Government. The U.S. Government is authorized to reproduce and distribute reprints for Government purposes notwithstanding any copyright notation here on.

{\small
\bibliographystyle{ieee_fullname}
\bibliography{egbib}
}

\end{document}